# Lexical Learning as an Online Optimal Experiment: Building Efficient Search Engines through Human-Machine Collaboration.

**Jacopo Tagliabue[†], Reuben Cohn-Gordon[‡]**

[†]Tooso Labs, San Francisco, CA, jacopo.tagliabue@tooso.ai
[‡]Dept. of Linguistics, Stanford University, Palo Alto, CA, reubencg@stanford.edu

## Abstract

Information retrieval (IR) systems need to constantly update their knowledge as target objects and user queries change over time. Due to the power-law nature of linguistic data, learning lexical concepts is a problem resisting standard machine learning approaches: while manual intervention is always possible, a more general and automated solution is desirable. In this work, we propose a novel end-to-end framework that models the interaction between a search engine and users as a virtuous human-in-the-loop inference. The proposed framework is the first to our knowledge combining ideas from psycholinguistics and experiment design to maximize efficiency in IR. We provide a brief overview of the main components and initial simulations in a toy world, showing how inference works end-to-end and discussing preliminary results and next steps.

## Introduction

Information retrieval (IR) systems play an important part in the digital life of billions of people daily and in the overall economy: taking fashion as an industry example, 25% of all transactions are now happening online (Statista 2019). Due to the interactive nature of IR, search engines are great use cases for studying human-machine collaboration "in the wild".

In *this* paper, we sketch a novel framework drawing from psycholinguistics and experiment design to address lexical learning in IR through human-in-the-loop inference: this end-to-end pipeline provides principled lexical learning to IR systems *without* manual intervention or large amount of data. As such, the proposed framework sits ideally in between human experts and machine learning. Human curated resources can be seamlessly used as *input* to the system, which then exploits the Bayesian apparatus to evolve autonomously over time. On the other hand, the system can supplement standard distributional approaches (Grbovic *et al.* 2016), helping precisely where they are notoriously less effective, i.e. the "long tail" (Bai *et al.* 2018)[1].

## Problem statement

Consider the stylized IR interface in Figure 1, showing how User and System take turns in discovering products on a digital store (in the example, User is looking for *ruffle dresses*):

a) User searches for objects using a word $z$;
b) the System realizes $z$ is unknown, and asks the User to pick a "close one" from a small set of objects;
c) the User gives feedback by clicking on at most one object.

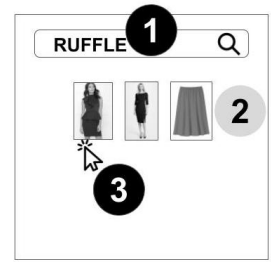

Figure 1: a stylized IR interface for so-called "product search".

Intuitively, the problem is the following: how can we design a System that learns the meaning of $z$ as quickly as possible, leveraging User's feedback? The answer requires us to solve two problems: first, System must be able to reason over meaning hypotheses for $z$ in a consistent fash-ion; second, System must be able to ask User the "right" question (in the form of product selection) to maximize learning.

---

[1] The problem statement also resembles topics in the bandit and active learning community (e.g. Bouneffouf et al. 2014): the main difference is that in the present case symbolic knowledge is both the source of meaning hypotheses *and* the engine for optimal experiments.

## Reasoning over meaning

Knowledge graphs (KGs) are spreading (Sicular and Brant, 2018) as an effective backbone for IR (He et al., 2016). If System represents products as a taxonomy (see Figure 2), given an unknown word $z$ and nodes as meaning hypotheses, the posterior probability of each node, given User's clicks $x_1, x_2, \ldots x_n$ in $X$, is given by Bayes formula: $P(node|X) \propto P(X|node) * P(node)$.

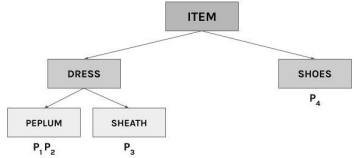

Figure 2: a stylized fashion KG with 4 products and 5 nodes.

Extending the intuitions of (Xu and Tenenbaum, 2007) to leverage KGs, we define a prior favoring "ontological distinctiveness"[2] (OD) for nodes as:

1) $P(node) \propto OD(node, sibling(node))$

Likelihood follows the "size principle": if $ext(node)$ is the function assigning products to a node, we define likelihood for a single observation $x_i$ as:

2) $P(x_i|node) \propto \delta \cdot (1/|ext(node)|) + bernoulli(\varepsilon)$

where $\delta = 1$ if $x_i \in ext(node)$, 0 otherwise. In other words, smaller hypotheses are preferred and a noise parameter $\varepsilon$ accounts for erratic user behavior (e.g. how much can we trust User picking $P_1$ to really mean that $P_1$ is a *peplum*?).

## Asking for human help efficiently

Consider again the KG in Figure 2: for bundle size $n=2$, System can show User $[P_2, P_1]$ vs $[P_3, P_2]$ vs $[P_4, P_3]$, etc. and then observe what she clicks. What System wants, given its prior over nodes, is to show bundle $b^*$ such that User action $y$ over $b^*$ maximally updates the distribution (Ouyang et al., 2016), as measured by the expected Kull-back–Leibler divergence. Since feedback is not known in advance, we need to marginalize over all the possible $y$, giving:

3) $b^* = \text{argmax}(b) \; E_{p(y;\, b)} \; D_{KL}(P(node) \,|\, b, y) \,\|\, P(node))$

If we calculate the expected information gain (EIG) for all bundles in Figure 2, $[P_4, P_3]$ is the best one[3].

## Putting all together in a virtuous loop

Going back to the original problem statement, it is now easy to see how System can leverage User to efficiently gain knowledge over an unknown word $z$: System will first prepare meaning hypotheses over $z$ and then calculate the product bundle maximizing EIG. Once User clicks on a product, the Bayesian inference will produce a posterior distribution which can be used as prior for a second interaction (and so on, in a virtuous loop). In the case of the KG from Figure 2, if $z=footwear$, $[P_4, P_3]$ will be the test bundle; after observing User clicking on $P_4$, System's confidence on the meaning of $z$ is already converging over the correct node *shoes* (~98% average confidence over 10 MCMC runs[4]).

It's crucial to note that such efficiency is the result of *combining* the two ideas and that neither is sufficient alone: KG-based lexical acquisition can work only by conditioning on reasonable feedback, such as the data provided by maximizing EIG[5]; on the other hand, selecting the perfect bundle would be worthless without an inferential model that is good at leveraging the structure of the hypothesis space.

## Conclusion

We presented a novel framework for IR systems and show preliminary results from simulations: both components are principled and independently motivated, and we look forward to explore the computational model in more realistic use cases (e.g. varying KGs, preferences, noise).

While many practitioners believe IR to be mostly a "Big Data" problem, the extreme power-law nature of language interfaces makes for a compelling "small data" business case[6]: there is great theoretical and practical value in developing alternative approaches, as human-machine collaboration promises to make optimal use of every search interaction, no matter how rare.

---

[2] If a given KG represents product features, a natural choice to measure OD would be calculating the Jacquard Distance between the feature set of the given nodes.

[3] Intuitively, this is correct since a single click on $P_4$ will make the node *shoes* much more likely.
[4] Please see the additional materials for the relevant charts.
[5] Note for example that if product bundle is chosen randomly, only ~50% of the time $P_4$ would be selected.
[6] Please see additional materials showing query distribution for real websites: it is not unusual that a vast amount of traffic is generated by rare queries (the "long tail" of the distribution), for which standard data hungry statistical tools may give inaccurate results.